\documentclass{article}
\pdfoutput=1
\usepackage[utf8]{inputenc} 
\usepackage[T1]{fontenc} 
\usepackage{hyperref} 
\usepackage{url} 
\usepackage{booktabs} 
\usepackage{amsfonts} 
\usepackage{nicefrac} 
\usepackage{microtype} 
\usepackage{graphicx} 
\usepackage{array}
\usepackage{amsmath}
\usepackage{placeins}
\usepackage{subcaption}
\usepackage{caption}
\usepackage[square,sort,comma,numbers]{natbib}
\usepackage{fancyhdr}
\usepackage{tabularx}
\usepackage{accents}
\usepackage{color}
\usepackage{multirow}

\usepackage[
a4paper,
textheight=9in,
textwidth=6in,
top=1in
]{geometry}
\renewenvironment{abstract}%
{%
\vskip 0.075in%
\centerline%
{\large\bf Abstract}%
\vspace{0.5ex}%
\begin{quote}%
}
{
\par%
\end{quote}%
\vskip 1ex%
}

\date{}

\title{Training Deep Neural Network in Limited Precision}

\author{
  Hyunsun Park\thanks{Authours contributed equally.}, 
  Jun Haeng Lee$^*$, Youngmin Oh, Sangwon Ha, Seungwon Lee  \\
  Samsung Advanced Institute of Technology \\
  Samsung-ro 130, Suwon-si, Republic of Korea \\
  \texttt{\{h-s.park, junhaeng2.lee\}@samsung.com} \\
}

\begin{document}

\maketitle

\begin{abstract}
Energy and resource efficient training of DNNs will greatly extend the applications of deep learning.
However, there are three major obstacles which mandate accurate calculation in high precision.
In this paper, we tackle two of them related to the loss of gradients during parameter update and backpropagation through a softmax nonlinearity layer in low precision training.
We implemented SGD with Kahan summation by employing an additional parameter to virtually extend the bit-width of the parameters for a reliable parameter update. 
We also proposed a simple guideline to help select the appropriate bit-width for the last FC layer followed by a softmax nonlinearity layer.
It determines the lower bound of the required bit-width based on the class size of the dataset.
Extensive experiments on various network architectures and benchmarks verifies the effectiveness of the proposed technique for low precision training.
\end{abstract}

\section{Introduction}
\label{section:intro}

Employing accelerators equipped with low precision computation elements can significantly improve the energy efficiency of operating state-of-the-art deep neural networks (\textbf{DNN}s) \cite{chen2017eyeriss, han2016eie, zhang2016cambricon, micikevicius2017mixed, das2018mixed}.
Although abundant previous works exist on converting or training DNNs for low precision inference \cite{courbariaux2015binaryconnect, li2016ternary, zhu2016trained, hubara2016binarized}, most of them require full precision hardware (\textbf{HW}) for training.

Enabling training as well as inference on the edge devices empowered with accelerators based on low precision computation units can open doors for many personalization applications previously restricted due to privacy issues \cite{mcmahan2017federated}.
For example, sensitive data such as biometrics can safely be consumed for training with in the trust zone of a personal device.
Nevertheless, computing in low precision is critical in order to reduce power consumption and memory footprint.

A few methods were proposed to train DNNs in limited precisions \cite{courbariaux2015traininglowprecision, gupta2015limited, miyashita2016logcnn, zhou2016dorefa, rastegari2016xnor}. 
However, the accuracies were not comparable to the full precision version.
Some of them even require special HW for unconventional operations, namely, log-scale calculation and/or stochastic quantization \cite{ miyashita2016logcnn, hubara2016binarized, hubara2016quantized, rastegari2016xnor, zhou2016dorefa}.
As mentioned in \cite{micikevicius2017mixed}, three problems have been commonly identified so far in training DNNs in limited precision: parameter update, softmax nonlinearity, and normalization.
In previous works, these were computed in full precision to avoid catastrophic accuracy degradation.

Recently, mixed precision HWs using both the low and high precision units together was proposed for accelerated training \cite{micikevicius2017mixed, koster2017flexpoint}. 
Speedup arises from reduced memory footprint and increased number of arithmatic units thanks to the low precision operations.
Massive portions of the calculations for the forward and backward propagations were accelerated by utilizing the low precision units while the high precision units were used to carefully handle the parameter update, softmax nonlinearity, and normalizations.
However, the accelerators must contain both low and high precision HW.

In this paper, we tackle two issues known to the parameter update and the softmax nonlinearity in low precision training. 
We propose to use Kahan summation \cite{kahan1965} in SGD for updating the network parameters with low precision computation units.
In addition, a simple method of selecting sufficient bit-width for the last fully-connected layer followed by a softmax layer based on the number of the output classes is proposed.

To investigate the effectiveness of the proposed methods, we selected an 8-bit accelerator supporting dynamic fixed-point with 8 bits or 16 bits for inference and training as the target HW model. 
16-bit operation is accomplished by using 8-bit computing units in multiple cycles without any additional HW unit. 
We kept the HW model as simple as possible by using uniform linear quantization without stochastic computing.
At any rate, the proposed methods can be applied to any type of number systems.

\section{SGD with the Kahan summation for low precision parameter update}
\label{section:lazyupdate}
Parameter set $\theta$ for a neural network can be optimized with the stochastic gradient descent (\textbf{SGD}) method \cite{bottou2010large}  as

\begin{equation}
\label{eq:sgd}
  \theta = \theta - \eta \frac{\partial L}{\partial \theta}
\end{equation}

where $L$ is the loss function, $\frac{\partial L}{\partial \theta}$ is the gradient of the loss in respect to the parameter at a given input batch, and $\eta$ is the learning rate.
In general, $\eta \frac{\partial L}{\partial \theta}$ is much smaller than $\theta$ by a few orders of magnitude.
When using limited numerical precision (e.g., 8 or 16 bits), $\eta \frac{\partial L}{\partial \theta}$ is often too small to change $\theta$ due to the lack of precision when $\theta$ is updated by (\ref{eq:sgd}).
Thus, the portion of the gradient vector used to update the parameter could significantly deviate from the original, which could substantially deteriorate training accuracy. 
To overcome this issue, we propose to use an extra parameter to maintain the partial accumulation of $\eta \frac{\partial L}{\partial \theta}$ and update the parameter only when the accumulated amount is large enough to be representable within the given precision and be able to actually change the parameter values.
The accumulation parameter acts as a carrier aggregating the small gradients and delivering them to the parameter.
The parameter update is accomplished in two stages through the accumulation parameter. 
This process is known as Kahan summation \cite{kahan1965}, which was proposed to reduce the numerical error in the total obtained by adding a sequence of finite precision numbers.
We call the SGD with the Kahan summation the \textbf{lazy update} due to the fact that the gradients are not likely to be applied to the parameters at every update step. 

\begin{figure}
  \centering
  \includegraphics[width=12cm]{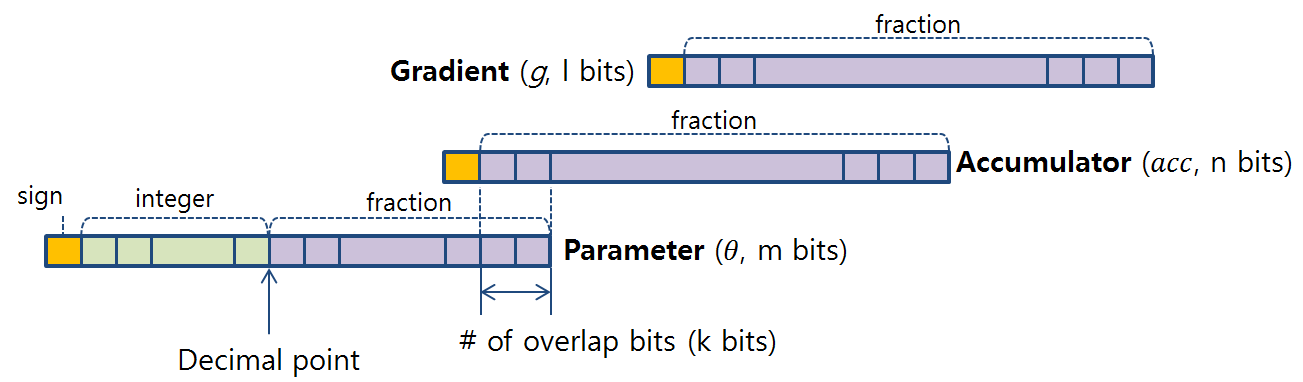}
  \caption{Illustration of the lazy update (SGD with the Kahan summation) in low precision dynamic fixed-point number system. Decimal points are aligned for parameter, gradient, and accumulator on the horizontal axis.
		Gradient value is too small to be applied to the parameter directly. 
		Accumulator accumulates gradients for multiple iterations until its value becomes large enough to be added to the parameter.}
  \label{fig:lazy_update}
\end{figure}

When the accumulation of $\eta \frac{\partial L}{\partial \theta}$ is large enough, we update $\theta$ with the value that is within the supported precision range only and preserve the remainder in the accumulation for later use.
Figure \ref{fig:lazy_update} shows the operating principle of the lazy update in a dynamic fixed-point number system.
The parameter, the gradient, and the accumulator all have the same decimal point location on the horizontal axis.
If the gradients are accumulated into an accumulator for multiple iterations until its values is large enough to change the targeted parameter value, only a few bits of the accumulator will overlap with the parameter. 
If $k$ bits overlap, the bit-width of the parameter can be effectively extended to ($m$ + $n$ – $2$ – $k$) bits excluding the sign bit, where $m$ and $n$ represent the number of bits for the parameter and the accumulator.
For example, if we use an 8-bit integer format (\textbf{INT8}) for the parameter and a 16-bit integer format (\textbf{INT16}) for the accumulator, we can provide a maximum of 22 bits for the parameter update, which is close to the bit-width of the fraction part of the 32-bit floating point format (\textbf{FP32}).

The lazy update can be implemented in a pure algorithmic way as shown in Algorithm 1 in low precision computing systems.
In addition, it can also be applied in a simpler manner on a specialized HW.
We emphasize that this idea can be applied to any numerical format: both floating-point and fixed-point representations. 

\begin{table}
\centering
  \label{table:lazyupdate_algo}
  \begin{tabular}{l}
    \toprule
    \textbf{Algorithm 1} Lazy update (SGD with the Kahan summation). \\ $acc$ is the accumulator for lazy update shown in Figure \ref{fig:lazy_update}.\\
    \midrule
    \textbf{Require:} $\theta$, $\eta \frac{\partial L}{\partial \theta}$, previous $\theta$, previous $acc$ \\
    \textbf{Ensure:} Update $\theta$, $acc$  \\ 
    \quad \quad \# Accumulate gradient. Due to finite precision, only effective value of $\eta \frac{\partial L}{\partial \theta}$ in the \\
    \quad \quad \# significant figure of $acc$ will be added to $\theta$ \\
    \quad \quad $acc \gets acc + \eta \frac{\partial L}{\partial \theta}$  \\

    \quad \quad \# Update $\theta$ with $acc$. Due to finite precision, only effective value of $acc$  in the \\
    \quad \quad \# significant figure of $\theta$ will be added to $\theta_{update}$ \\
    \quad \quad $\theta_{update} = \theta - acc$  \\

    \quad \quad \# Update residual value by subtracting updated portion.  \\
    \quad \quad \# ($\theta_{update} - \theta$) is different from $acc$ due to the finite precision. \\
    \quad \quad $acc \gets acc + (\theta_{update} - \theta)$  \\

    \quad \quad \# Update parameter.  \\
    \quad \quad $\theta \gets \theta_{update}$ \\

    \bottomrule
  \end{tabular}
\end{table}

\section{Classifier with softmax and cross entropy loss}
  \label{section:softmax}
In previous works on low precision training, the last fully-connected (\textbf{FC}) layer was usually trained in full precision since the required precision for the last FC layer showed complicated behavior depending on the benchmark datasets \cite{courbariaux2015binaryconnect, micikevicius2017mixed, zhou2016dorefa}.
However, there has been no clear explanation on such a behavior. 
In this section, we explain why the last FC layer requires benchmark-dependent precision unlike other layers and propose a simple way to select an appropriate bit-width. 

\begin{figure}
  \centering
  \begin{subfigure}[b]{0.2\textwidth}
    \centering
     \includegraphics[width=\textwidth]{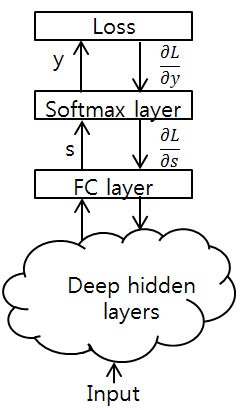}
     \caption{ \label{fig:sub_softmax_1}}
  \end{subfigure}
  \quad  \quad
  \begin{subfigure}[b]{0.65\textwidth}
     \centering
     \includegraphics[width=\textwidth]{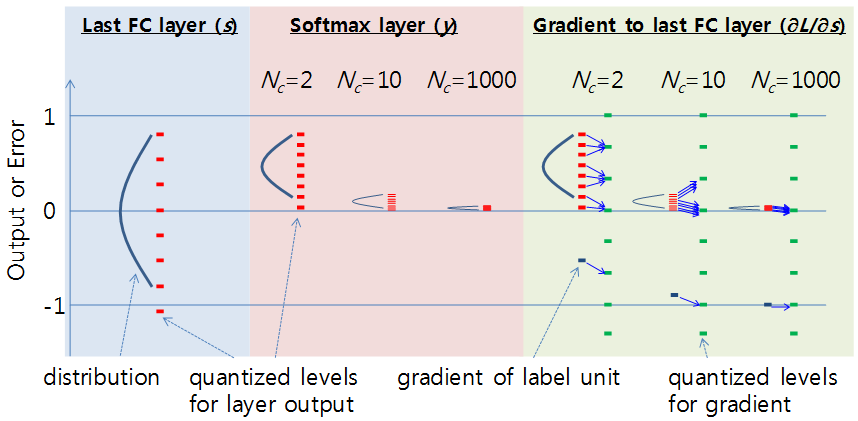}
     \caption{ \label{fig:sub_softmax_2}}
  \end{subfigure}
  \caption{(a) Typical deep neural network architecture for classification task. (b) Example distribution of the output of the last FC layer ($s$, left), the output of softmax layer ($y$, center), and the gradient delivered from the softmax layer to the FC layer ($\partial L / \partial s$, right) at an early stage of training, and quantization levels for them in dynamic INT3. $N_c$ is the class size which is equal to the number of units in the last FC layer. As $N_c$ increases, most output values of the softmax layer are packed close to zero due to normalization. Thus, most elements of $\partial L / \partial s$ become zero by low precision quantization.}
  \label{fig:softmax}
\end{figure}

The last FC layer (i.e. classifier) is commonly followed by a softmax nonlinearity layer with the cross entropy loss for training in a network for a classification task as shown in Figure \ref{fig:sub_softmax_1}.
The softmax layer converts the output of the last FC layer to the inferred probability of each class. 

The cross entropy loss of the $i$-th output is defined as

\begin{equation}
\label{eq:softmax}
  L = -\sum_{i=1}^{N_c} t_i log(y_i), \; \mathrm{where} \; y_i = \frac{e^{s_i}}{\sum_{c=1}^{N_c} e^{s_c}}
\end{equation}

Here, $N_c$ is the class size (i.e. the number of classes), $s_i$, $y_i$, and $t_i$ are the $i$-th output of the last FC layer, the output of the softmax layer, and the ground truth (\textbf{GT}) label, respectively. 

The gradient of the cross entropy loss in (\ref{eq:softmax}) is calculated as

\begin{equation}
\label{eq:gradient}
  \frac{\partial L}{\partial s_i} = \sum_{k=1}^{N_c} \frac{\partial L}{\partial y_k} \frac{\partial y_k}{\partial s_i} = y_i - t_i
\end{equation}

Figure \ref{fig:sub_softmax_2} shows the distribution of $s$, $y$, and $\partial L / \partial s$ at an early stage of training and their quantized levels. 
Since the network is at an early stage of training, the output of the last FC layer ($s$ in the left Figure \ref{fig:sub_softmax_2}) is distributed around zero.
On the other hand, the output of the softmax layer ($y$ in the center Figure \ref{fig:sub_softmax_2}) is concentrated on the level 1/$N_c$ since its output values are normalized so that they sum up to one (i.e. $\sum_{c=1}^{N_c}y_i =1$).
For example, when $N_c$ is 1000, all the values are around 0.001.
Applying (\ref{eq:gradient}) to this case makes the gradient vector $\partial L / \partial s$ have a value close to -1 for the element with the GT label and values around 0.001 at all others which, unfortunately, will be truncated to be zero with low precision quantization (shown in the right Figure \ref{fig:sub_softmax_2}).
We realized that this one-hot-vector-like gradient tends to make training unstable. 
The sum of all the elements in the gradient vector is considerably biased to a negative value since most of the small positive errors vanish due to quantization.
Those small values of the gradient vector play an important role in training as stated in \cite{hinton2015distilation}. 
The magnitude of the small components is inversely proportional to the class size.
Therefore, we propose to determine the bit-width for the last FC layer based on the class size. 

The results in Section \ref{section:small_networks} show that the simple tasks with 10 classes can be trained with dynamic INT8 for all layers.
However, INT16 was required for the last FC layer to train networks on the 1000-class ImageNet benchmark even though INT8 worked for all the other layers.
If we want to keep the sum of round-off components of the gradient vector $\partial L/\partial s$ less than $\alpha  (0<\alpha<1)$, the following condition must be satisfied for dynamic fixed-point with linear quantization:

\begin{equation}
\label{eq:class}
 \frac{\alpha}{\mathrm|class| -1 } \geq  {2}^{\mathrm-(BW_{classifier}-1)}
\end{equation}

 where |class| is the class size. 

Thus, we can derive a simple rule of thumb for the required bit-width for the last FC layer followed by the softmax and cross entropy loss layers as follows:

\begin{equation}
\label{eq:bit_width}
  \mathrm{bitwidth_{classifier}} > log_2(\mathrm{|class|-1}) + log_2(\frac{2}{\alpha})
\end{equation}

We empirically found that $\alpha = 0.5$ is a reasonable choice.

\section{Benchmark results}
\label{section:results}
We evaluated our method on a variety of neural networks that include convolutional neural networks (\textbf{CNN}s) for image classification, general adversarial networks (\textbf{GAN}s) for image generation, and  recurrent neural network (\textbf{RNN}) based on long-short-term memory (\textbf{LSTM}) units for speech recognition.
We modified Caffe\cite{jia2014caffe} and Tensorflow\cite{abadi2016tensorflow} to support the dynamic fixed-point data formats for training.

The lazy update described in Algorithm 1 was implemented in the optimizers. 
We found that INT16 was necessary for the ADAM optimizer\cite{kingma2014adam} because of $\beta_2$ (= 0.999 normally) while the momentum optimizer will still work in INT8.
For comparison purposes, the same hyperparameters were used as those from training the networks in full precision (i.e., FP32) except where otherwise noted.

\subsection{CNNs on MNIST, SVHN and CIFAR10}
\label{section:small_networks}

This section demonstrates the training results for four small-scale CNNs on MNIST\cite{lecun2010mnist}, SVHN\cite{netzer2011reading} and CIFAR10\cite{krizhevsky2009learning} datasets. 
Both MNIST and SVHN are digit image datasets, totalling 10 classes from 0 to 9. MNIST consists of handwritten digit images, which has 60,000 training images and 10,000 testing images. 
SVHN is a real-world dataset and contains 73,257 digit images for training and 26,032 samples for testing. 
CIFAR10 is composed of 10 classes for $32\times32$ color images. There are 50,000 training images and 10,000 testing images.

All networks were configured with multiple convolutional (\textbf{CONV}) modules consisting of convolution, ReLU activation, and pooling layers. 
We made variations in the network architecture by adding batch normalization\cite{ioffe2015batch} (\textbf{BN}) layers and skipping connections to investigate the proposed method over various architectures.
We trained LeNet\cite{lecun1998gradient} consisting of two CONV modules and two FC layers on the MNIST dataset. 
The network for the SVHN benchmark has four CONV modules and  two FC layers. 
CIFAR10 was trained with two networks; one of which has three CONV modules followed by one FC layer (\textbf{CIFAR10-CNN}) and the other ResNet20 (\textbf{CIFAR10-ResNet20})\cite{he2016deep}.
The CNN for SVHN and RestNet20 for CIFAR10 contained BN layers. 
We also investigated two optimizer types: momentum SGD and ADAM. 
The CNN for SVHN was trained using the ADAM optimizer ($\beta_1$ = 0.9, $\beta_2$ = 0.999) while the momentum optimizer ($\mu$ = 0.9) was used for other networks.
The inputs and outputs of BN layers were also quantized, but internal parameters and operations were done without quantization to avoid instability during training.

\begin{figure}
  \centering
  \begin{subfigure}[b]{0.405\textwidth}
    \centering
     \includegraphics[width=\textwidth]{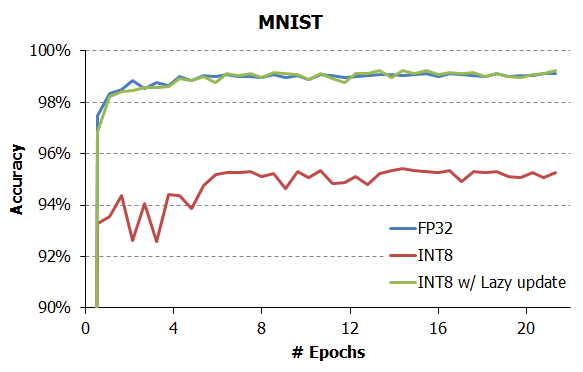}
     \caption{MNIST \label{fig:sub_mnist}}
  \end{subfigure}
  \quad  \quad
  \begin{subfigure}[b]{0.405\textwidth}
     \centering
     \includegraphics[width=\textwidth]{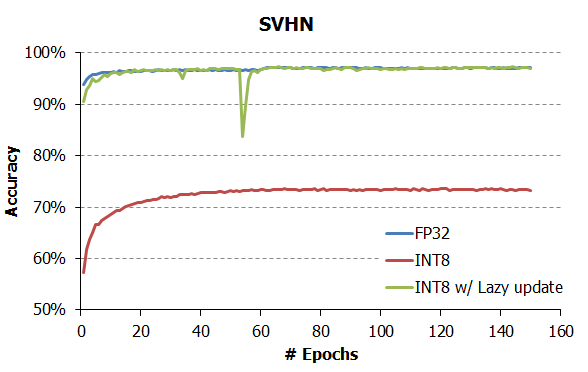}
     \caption{SVHN \label{fig:sub_svhn}}
  \end{subfigure}
 \vskip\baselineskip
  \begin{subfigure}[b]{0.405\textwidth}
     \centering
     \includegraphics[width=\textwidth]{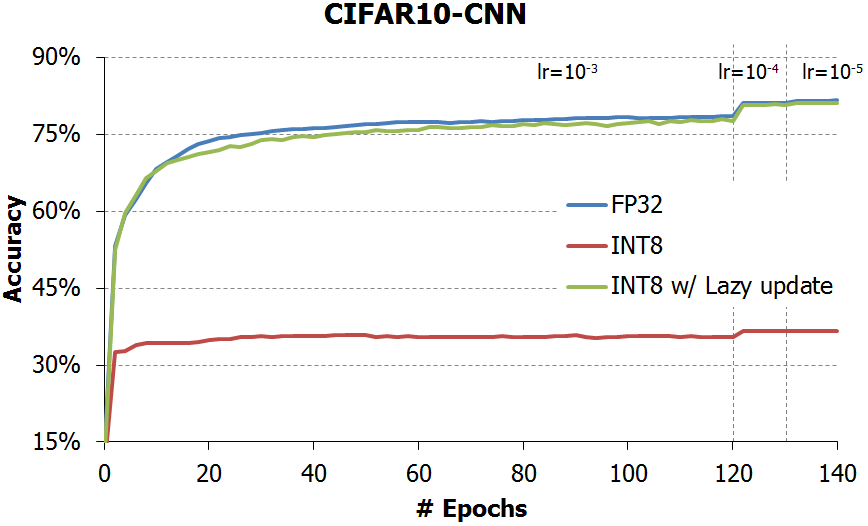}
     \caption{CIFAR10-CNN \label{fig:sub_cifar10_conv}}
  \end{subfigure}
  \quad  \quad
  \begin{subfigure}[b]{0.405\textwidth}
     \centering
     \includegraphics[width=\textwidth]{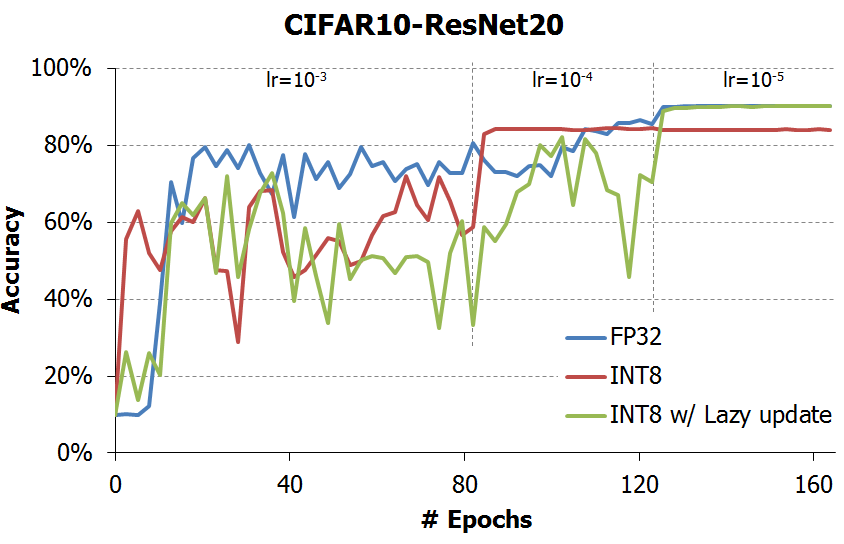}
     \caption{CIFAR10-ResNet20 \label{fig:sub_cifar10_res}}
  \end{subfigure}
  \caption{Convergence plots for FP32, INT8 and our method when training for MNIST, SVHN, and CIFAR10.\label{fig:convergence}}
\end{figure}

Figure \ref{fig:convergence} shows the convergence of the validation accuracy measured on the benchmarks during training.  
We observed a huge accuracy degradation (up to 44.9\%) in all cases when we trained the networks in INT8 due to the limited precision during the parameter update.  
The effect of limited precision is clearly shown in CIFAR10-ResNet20 where the accuracy did not improve in INT8 unlike FP32 where the learning rate was reduced at 120 epochs. 
However, when we added the lazy update in the optimizers, the accuracy converged to a level comparable with FP32 in all cases. 
The lazy update could effectively aggregate small weight gradients to update the parameters.  
Table \ref{table:small_result} summaries the final validation accuracies for FP32, INT8, and our method (INT8 with the lazy update). 

\begin{table}
  \caption{Validation accuracies of CNNs on MNIST, SVHN, and CIFAR10 datasets .\label{table:small_result}}
  \centering
  \begin{tabular}{lclclcl}
    \toprule
    Benchmark & FP32 & INT8 & INT8 with lazy update \\
    \midrule
    MNIST 			& 99.10\% & 95.28\% & 99.24\% \\   
    SVHN			& 97.06\% & 73.32\% & 96.99\% \\   
    CIFAR10-CNN 	& 81.56\% & 36.66\% & 81.17\% \\   
    CIFAR10-ResNet20 	& 90.16\% & 83.84\% & 90.23\% \\   
    \bottomrule
  \end{tabular}
\end{table}

\subsection{AlexNet on ImageNet dataset}
In this section, we evaluate the lazy update on a large scale network on the ImageNet dataset\cite{ILSVRC15}. 
We trained AlexNet\cite{krizhevsky2012imagenet} using the momentum optimizer. 
As mentioned in Section \ref{section:softmax}, we used INT16 for the weight parameters, the activation and the gradient values between the last FC layer and the softmax layer to accommodate 1,000 classes of the ImageNet benchmark. 
From (\ref{eq:bit_width}), the bit-width needs to be larger than 14 bits (13.97 = $log_2(1000) + 4$). 
We also used INT16 for the momentum optimizer ($\mu$ = 0.9).
The learning rate was initialized to 0.01 and reduced by a factor of 5 every 100k updates.

As expected, straight-forward training of AlexNet in normal INT8 failed.
The accuracy did not improve at all from the chance probability (0.001).
It was possible to train the network in INT16 with some accuracy degradation (2.82\%), but we were able to obtain a better accuracy by adding the lazy update in INT8 as shown in Table \ref{table:alexnet_result}.
The lazy update using INT8 led to only 0.25\% accuracy loss compared with FP32 training. 

\begin{figure}
  \centering
  \includegraphics[width=0.6\textwidth]{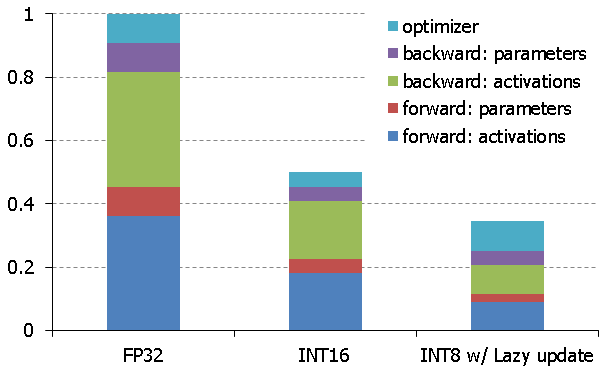}
  \caption{Memory usage of AlexNet training with 256 batchs. Normalized FP32.\label{fig:alexnet_mem}}
\end{figure}

In order to investigate the advantage of our method in resource perspective, we calculated the memory usage required for training AlexNet with a mini-batch of 256 samples as shown in Figure \ref{fig:alexnet_mem}. 
Using INT16 obviously halved the memory usage compared to FP32.
INT8 further reduces it to half.
However, there is an increased memory usage to implement the lazy update in the optimizers. 
Our method of (INT8 + lazy update) can achieve 31.2\% reduction of the memory usage in addition to the improved accuracy compared to INT16.

\begin{table}
  \caption{Validation accuracies of AlexNet on ImageNet classification task.\label{table:alexnet_result}}
  \centering
  \begin{tabular}{lclclclc|clclc}
    \toprule
    \multicolumn{1}{c}{\multirow{2}{*}{Model}} & \multicolumn{2}{c}{FP32}& \multicolumn{2}{c}{INT16}& \multicolumn{2}{c}{INT8 with lazy update }\\ 
    \cmidrule(r){2-7}
    \multicolumn{1}{c}{}	& top-1 & top-5 & top-1 & top-5 & top-1 & top-5 \\
    \midrule
    AlexNet 		& 58.17\% & 80.81\% & 55.35\% & 78.69\% & 57.92\% & 80.62\%\\    
    \bottomrule
  \end{tabular}
\end{table}

\subsection{Transfer learning on sub-problems of ImageNet dataset}
Transfer learning will be one of the major applications of low-precision training on edge devices. In this section, we evaluate our method on transfer learning scenarios. 
We made two small datasets from ImageNet dataset by grouping classes into super classes: hunting dog and machine. 
Each dataset is composed of 58 classes and 44 classes, respectively. 
Since we use pre-trained models on ImageNet training dataset, we constructed the new datasets from the ImageNet validation dataset to avoid using the same image samples during the transfer learning. 
The ImageNet validation dataset has 50 images for each class, of which, 45 images are used for training and 5 images are used for validation in our experiments. 
Pre-trained AlexNet and Inception-v3 [36] are used as base architectures. 
We replace the classifier (i.e. last FC layer) to adapt to the new classification tasks. 
Since the class size is small, we used INT8 for the classifier. 

Table \ref{table:transfer_result} summarizes the results of the transfer learning experiment. 
Training Incpetion-v3 with INT8 results in accuracy loss up to -4.14\% on hunting dog dataset. 
However, with the lazy update, there is no accuracy degradation. 
INT16 also shows comparable accuracy, but it requires more memory and longer operation duration.

\begin{table}
  \caption{Validation accuracies of transfer learning on small ImageNet dataset.\label{table:transfer_result}}
  \centering
  \begin{tabular}{lclclclc|c|c}
    \toprule
    \multicolumn{2}{c}{Benchmark} & FP32 & INT16 & INT8 & INT8 with lazy update \\
    \midrule
    \multirow{2}{*}{AlexNet}   	& Hunting dog 	& 65.86\% & 65.86\% &64.48\% & 66.90\% \\   
    \multicolumn{1}{c}{}		&Machine 	& 77.27\% & 78.18\% &77.27\% & 77.73\% \\    
    \multirow{2}{*}{Incption-v3}  & Hunting dog 	& 86.55\% & 86.21\% &82.41\% & 86.55\% \\   
    \multicolumn{1}{c}{}   		& Machine 	& 91.82\% & 91.82\% &90.91\% & 91.82\% \\    
    \bottomrule
  \end{tabular}
\end{table}

\subsection{Image generation with GANs}

We applied the lazy update scheme to train GANs in low precision fixed point formats.\
A GAN consists of two neural networks competing against each other. 
One of them (the generator) generates fake images and tries to deceive the other (the discriminator) which distinguishes whether the input images are real or not. 

We trained BEGAN\cite{berthelot2017began} and DCGAN\cite{radford2015unsupervised} on the CelebA\cite{liu2015faceattributes} dataset with cropping and alignment and the LSUN bedrooms\cite{yu15lsun} dataset, respectively, which are representative datasets in the GAN experiments. 
The CelebA dataset consists of $202,599$ facial images and the LSUN bedrooms dataset consists of $3,033,042$ images for training and 300 images for validation.
The network parameters were quantized in INT8.
However, INT16 was required for the activations to generate good quality images. 
The ADAM optimizer was used in INT16 with $\beta_1$ = 0.5, $\beta_2$ = 0.9, $\epsilon$ = $10^{-3}$, and the learning rate $10^{-4}$. 

\begin{figure}
  \centering
  \begin{subfigure}[b]{0.45\textwidth}
    \centering
     \includegraphics[width=\textwidth]{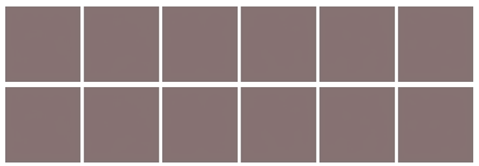}
     \caption{BEGAN INT8 \label{fig:began_int8}}
  \end{subfigure}
  \quad  \quad
  \begin{subfigure}[b]{0.45\textwidth}
     \centering
     \includegraphics[width=\textwidth]{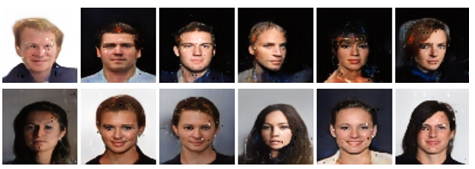}
     \caption{BEGAN INT8 with lazy update \label{fig:began_lazyupdate}}
  \end{subfigure}
 \vskip\baselineskip
  \begin{subfigure}[b]{0.45\textwidth}
     \centering
     \includegraphics[width=\textwidth]{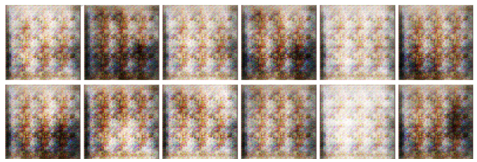}
     \caption{DCGAN INT8 \label{fig:dcgan_int8}}
  \end{subfigure}
  \quad  \quad
  \begin{subfigure}[b]{0.45\textwidth}
     \centering
     \includegraphics[width=\textwidth]{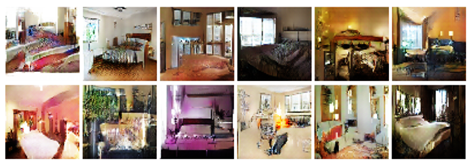}
     \caption{DCGAN INT8 with lazy update \label{fig:dcgan_lazyupdate}}
  \end{subfigure}

  \caption{All GANs above generated $64\times64$ faces and bedrooms images with INT8
weights and INT16 activations, gradients and optimizers in all layers.\label{fig:GANs}}
\end{figure}

Figure \ref{fig:GANs} illustrates the images generated by BEGAN and DCGAN.
The effectiveness of the lazy update is clearly shown in Figure \ref{fig:began_int8} and \ref{fig:began_lazyupdate}. 
BEGAN trained in INT8  failed to produce faces all together (Figure \ref{fig:began_int8}) while applying the lazy update created faces which were indistinguishable with those trained in full precision (Figure \ref{fig:began_lazyupdate}).
We observed the same tendency for the DCGAN on the bedrooms dataset as shown in Figure \ref{fig:dcgan_int8} and \ref{fig:dcgan_lazyupdate}. 

\subsection{LSTM on TIDIGIT dataset}
To investigate the effectiveness of the lazy update on training RNNs in low precision fixed point formats, a simple LSTM network was trained on the TIDIGITS dataset  \cite{leonard1993tidigits}.  
The TIDIGITS dataset, which is a set of spoken digits (``zero'' to ``nine'' plus ``oh'') for classification, has 2,464 digit samples in the training set and 2,486 samples for the test set.
Individual digits were transformed to produce a 39-dimensional Mel-Frequency Cepstral Coefficient (MFCC) feature vector using a 25\,ms window, 10\,ms frame shift, and 20 filter bank channels.  
The labels for ``oh'' and ``zero'' were collapsed to a single label.  
The network has a single LSTM layer consisting of 200 units.
The final state of the LSTM layer was fed to an FC layer with 200 units followed by a classification layer for the 10 digit classes. 
We used the ADAM optimizer ($\beta_1$=0.9, $\beta_2$=0.999) for training.

\begin{figure}
  \centering
  \includegraphics[width=0.6\textwidth]{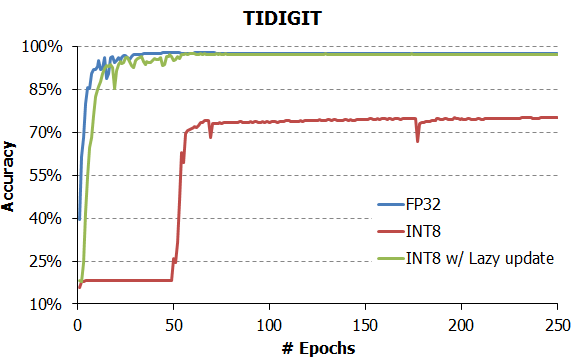}
  \caption{Convergence plot for TIDIGIT.\label{fig:tidigit}}
\end{figure}

Figure \ref{fig:tidigit} shows the validation accuracy measured during training on the TIDIGIT dataset. 
Training in INT8 failed to converge to the accuracy of FP32.
However, the accuracy loss improved from 22.41\% to 0.21\%, a level competitive to full precision by using the lazy update in the optimizer (see Table \ref{table:lstm_result}).

\begin{table}
  \caption{Validation accuracies of LSTM network on TIDIGIT dataset.\label{table:lstm_result}}
  \centering
  \begin{tabular}{lclclclc}
    \toprule
    Benchmark & FP32 & INT8 & INT8 with lazy update \\
    \midrule
    TIDIGIT 			& 97.62\% & 75.21\% & 97.41\% \\   
    \bottomrule
  \end{tabular}
\end{table}

\section{Related works}
\label{others}
There has been much effort to train deep neural networks (DNNs) operating in low precision for efficient inference\cite{courbariaux2015binaryconnect, hubara2016binarized, rastegari2016xnor, zhou2016dorefa}. 
They mainly focused on obtaining the networks with the weights or the activations in low precision. 
Other tensors for training like the gradient vectors required full precision data format.
Thus, training was done in hardwares with full precision.
For example, the Binaryconnect\cite{courbariaux2015binaryconnect} proposed training for binary weights. 
BNN\cite{hubara2016binarized} extended applying binarization to the activations. 
In QNN\cite{hubara2016quantized}, they tried to increase the bit-width of activations to 2, 4, and 6 bits in order to improve the training accuracy.
However, in all of these works, training was done in full precision.

XNOR-net\cite{rastegari2016xnor} attempted to binarize all the tensors including the gradients. 
Along these lines, most of the multiplications were substituted by additions in both forward and backward propagation. 
DoReFa-Net\cite{zhou2016dorefa} also proposed a method for training DNNs with quantized weights, activations, and gradients. 
This works used a different bit-width for each tensor to further improve accuracy. 
XNOR-net \cite{rastegari2016xnor} and DoReFa-Net\cite{zhou2016dorefa} quantized the activation gradients, but they used full precision for the weight gradient and kept the master parameters in FP32 for update. 

Flexpoint\cite{koster2017flexpoint} introduced a way to implement the dynamic fixed point for training.
It estimates the exponent value based on the history of the maximum values to prevent overflow. 
It achieved a similar accuracy with FP32 on various tasks. 
However, this numerical format requires specialized hardware and complicated calculation for the exponent value.  
Mixed precision training in \cite{das2018mixed} solves such overhead by scanning the maximum value of each tensor to determine the exponent value. 
However, they used FP32 accumulator to prevent overflow during accumulation of the results of fixed point multiplications. 
Mixed precision method in \cite{micikevicius2017mixed} proposed to use both half precision floating point (\textbf{FP16}) and FP32 for training.
The main datapaths for forward and backward propagations were accelerated by using FP16 while the parameter update, softmax nonlinearity, and normalizations were done in FP32.

\section{Conclusion}
In this paper, we addressed two major problems in low precision training.
We proposed the lazy update to avoid the problem caused by precision shortage in parameter update of DNNs on a HW with limited precision computation units. 
The lazy update employs an additional parameter to keep the partial accumulation of small gradient values for reliable update of parameters during training.
We can train DNNs in various network architectures (CNN, GAN, and LSTM) from scratch on various benchmarks (MNIST, SVHN, CIFAR10, ImageNet, CelebA, LSUN bedrooms, and TIDIGIT datasets) without accuracy degradation.

We also proposed a simple guideline to help select the appropriate bit-width for the last FC layer followed by a softmax nonlinearity layer.
It determines the lower bound of the required bit-width based on the class size of the benchmark.
Further study is required to alleviate the requirement of high precision computation in normalization layers and move toward energy efficient training of DNNs.

\small
\bibliographystyle{abbrv}
\bibliography{refs}

\begin{thebibliography}{10}

\bibitem{abadi2016tensorflow}
M.~Abadi, P.~Barham, J.~Chen, Z.~Chen, A.~Davis, J.~Dean, M.~Devin,
  S.~Ghemawat, G.~Irving, M.~Isard, et~al.
\newblock Tensorflow: A system for large-scale machine learning.
\newblock In {\em OSDI}, volume~16, pages 265--283, 2016.

\bibitem{berthelot2017began}
D.~Berthelot, T.~Schumm, and L.~Metz.
\newblock Began: Boundary equilibrium generative adversarial networks.
\newblock {\em arXiv preprint arXiv:1703.10717}, 2017.

\bibitem{bottou2010large}
L.~Bottou.
\newblock Large-scale machine learning with stochastic gradient descent.
\newblock In {\em Proceedings of COMPSTAT'2010}, pages 177--186. Springer,
  2010.

\bibitem{chen2017eyeriss}
Y.-H. Chen, T.~Krishna, J.~S. Emer, and V.~Sze.
\newblock Eyeriss: An energy-efficient reconfigurable accelerator for deep
  convolutional neural networks.
\newblock {\em IEEE Journal of Solid-State Circuits}, 52(1):127--138, 2017.

\bibitem{courbariaux2015binaryconnect}
M.~Courbariaux, Y.~Bengio, and J.-P. David.
\newblock Binaryconnect: Training deep neural networks with binary weights
  during propagations.
\newblock In {\em Advances in neural information processing systems}, pages
  3123--3131, 2015.

\bibitem{courbariaux2015traininglowprecision}
M.~Courbariaux, J.-P. David, and Y.~Bengio.
\newblock Training deep neural networks with low precision multiplications.
\newblock {\em arXiv preprint arXiv:1412.7024v4}, 2015.

\bibitem{das2018mixed}
D.~Das, N.~Mellempudi, D.~Mudigere, D.~Kalamkar, S.~Avancha, K.~Banerjee,
  S.~Sridharan, K.~Vaidyanathan, B.~Kaul, E.~Georganas, et~al.
\newblock Mixed precision training of convolutional neural networks using
  integer operations.
\newblock {\em arXiv preprint arXiv:1802.00930}, 2018.

\bibitem{gupta2015limited}
S.~Gupta, A.~Agrawal, K.~Gopalakrishnan, and P.~Narayanan.
\newblock Deep learning with limited numerical precision.
\newblock {\em arXiv preprint arXiv:1502.02551}, 2015.

\bibitem{han2016eie}
S.~Han, X.~Liu, H.~Mao, J.~Pu, A.~Pedram, M.~A. Horowitz, and W.~J. Dally.
\newblock Eie: efficient inference engine on compressed deep neural network.
\newblock In {\em Computer Architecture (ISCA), 2016 ACM/IEEE 43rd Annual
  International Symposium on}, pages 243--254. IEEE, 2016.

\bibitem{he2016deep}
K.~He, X.~Zhang, S.~Ren, and J.~Sun.
\newblock Deep residual learning for image recognition.
\newblock In {\em Proceedings of the IEEE conference on computer vision and
  pattern recognition}, pages 770--778, 2016.

\bibitem{hinton2015distilation}
G.~Hinton, O.~Vinyals, and J.~Dean.
\newblock Distilling the knowledge in a neural network.
\newblock {\em arXiv preprint arXiv:1503.0253}, 2015.

\bibitem{hubara2016binarized}
I.~Hubara, M.~Courbariaux, D.~Soudry, R.~El-Yaniv, and Y.~Bengio.
\newblock Binarized neural networks.
\newblock In {\em Advances in neural information processing systems}, pages
  4107--4115, 2016.

\bibitem{hubara2016quantized}
I.~Hubara, M.~Courbariaux, D.~Soudry, R.~El-Yaniv, and Y.~Bengio.
\newblock Quantized neural networks: Training neural networks with low
  precision weights and activations.
\newblock {\em arXiv preprint arXiv:1609.07061}, 2016.

\bibitem{ioffe2015batch}
S.~Ioffe and C.~Szegedy.
\newblock Batch normalization: Accelerating deep network training by reducing
  internal covariate shift.
\newblock {\em arXiv preprint arXiv:1502.03167}, 2015.

\bibitem{jia2014caffe}
Y.~Jia, E.~Shelhamer, J.~Donahue, S.~Karayev, J.~Long, R.~Girshick,
  S.~Guadarrama, and T.~Darrell.
\newblock Caffe: Convolutional architecture for fast feature embedding.
\newblock {\em arXiv preprint arXiv:1408.5093}, 2014.

\bibitem{kahan1965}
W.~Kahan.
\newblock Further remarks on reducing truncation errors.
\newblock {\em Communications of the ACM}, 8(1):40, January 1965.

\bibitem{kingma2014adam}
D.~P. Kingma and J.~Ba.
\newblock Adam: A method for stochastic optimization.
\newblock {\em arXiv preprint arXiv:1412.6980}, 2014.

\bibitem{koster2017flexpoint}
U.~K{\"o}ster, T.~Webb, X.~Wang, M.~Nassar, A.~K. Bansal, W.~Constable,
  O.~Elibol, S.~Gray, S.~Hall, L.~Hornof, et~al.
\newblock Flexpoint: An adaptive numerical format for efficient training of
  deep neural networks.
\newblock In {\em Advances in Neural Information Processing Systems}, pages
  1742--1752, 2017.

\bibitem{krizhevsky2009learning}
A.~Krizhevsky and G.~Hinton.
\newblock Learning multiple layers of features from tiny images.
\newblock 2009.

\bibitem{krizhevsky2012imagenet}
A.~Krizhevsky, I.~Sutskever, and G.~E. Hinton.
\newblock Imagenet classification with deep convolutional neural networks.
\newblock In {\em Advances in neural information processing systems}, pages
  1097--1105, 2012.

\bibitem{lecun1998gradient}
Y.~LeCun, L.~Bottou, Y.~Bengio, and P.~Haffner.
\newblock Gradient-based learning applied to document recognition.
\newblock {\em Proceedings of the IEEE}, 86(11):2278--2324, 1998.

\bibitem{lecun2010mnist}
Y.~LeCun, C.~Cortes, and C.~Burges.
\newblock Mnist handwritten digit database.
\newblock {\em AT\&T Labs [Online]. Available: http://yann. lecun.
  com/exdb/mnist}, 2, 2010.

\bibitem{leonard1993tidigits}
R.~G. Leonard and G.~Doddington.
\newblock Tidigits.
\newblock {\em Linguistic Data Consortium, Philadelphia}, 1993.

\bibitem{li2016ternary}
F.~Li, B.~Zhang, and B.~Liu.
\newblock Ternary weight networks.
\newblock {\em arXiv preprint arXiv:1605.04711}, 2016.

\bibitem{liu2015faceattributes}
Z.~Liu, P.~Luo, X.~Wang, and X.~Tang.
\newblock Deep learning face attributes in the wild.
\newblock In {\em Proceedings of International Conference on Computer Vision
  (ICCV)}, 2015.

\bibitem{mcmahan2017federated}
B.~McMahan and D.~Ramage.
\newblock Federated learning: Collaborative machine learning without
  centralized training data.
\newblock {\em Google Research Blog}, 2017.

\bibitem{micikevicius2017mixed}
P.~Micikevicius, S.~Narang, J.~Alben, G.~Diamos, E.~Elsen, D.~Garcia,
  B.~Ginsburg, M.~Houston, O.~Kuchaev, G.~Venkatesh, et~al.
\newblock Mixed precision training.
\newblock {\em arXiv preprint arXiv:1710.03740}, 2017.

\bibitem{miyashita2016logcnn}
D.~Miyashita, E.~H. Lee, and B.~Murmann.
\newblock Convolutional neural networks using logarithmic data representation.
\newblock {\em arXiv preprint arXiv:1603.01025v2}, 2016.

\bibitem{netzer2011reading}
Y.~Netzer, T.~Wang, A.~Coates, A.~Bissacco, B.~Wu, and A.~Y. Ng.
\newblock Reading digits in natural images with unsupervised feature learning.
\newblock In {\em NIPS workshop on deep learning and unsupervised feature
  learning}, volume 2011, page~5, 2011.

\bibitem{radford2015unsupervised}
A.~Radford, L.~Metz, and S.~Chintala.
\newblock Unsupervised representation learning with deep convolutional
  generative adversarial networks.
\newblock {\em arXiv preprint arXiv:1511.06434}, 2015.

\bibitem{rastegari2016xnor}
M.~Rastegari, V.~Ordonez, J.~Redmon, and A.~Farhadi.
\newblock Xnor-net: Imagenet classification using binary convolutional neural
  networks.
\newblock In {\em European Conference on Computer Vision}, pages 525--542.
  Springer, 2016.

\bibitem{ILSVRC15}
O.~Russakovsky, J.~Deng, H.~Su, J.~Krause, S.~Satheesh, S.~Ma, Z.~Huang,
  A.~Karpathy, A.~Khosla, M.~Bernstein, A.~C. Berg, and L.~Fei-Fei.
\newblock {ImageNet Large Scale Visual Recognition Challenge}.
\newblock {\em International Journal of Computer Vision (IJCV)},
  115(3):211--252, 2015.

\bibitem{yu15lsun}
F.~Yu, Y.~Zhang, S.~Song, A.~Seff, and J.~Xiao.
\newblock Lsun: Construction of a large-scale image dataset using deep learning
  with humans in the loop.
\newblock {\em arXiv preprint arXiv:1506.03365}, 2015.

\bibitem{zhang2016cambricon}
S.~Zhang, Z.~Du, L.~Zhang, H.~Lan, S.~Liu, L.~Li, Q.~Guo, T.~Chen, and Y.~Chen.
\newblock Cambricon-x: An accelerator for sparse neural networks.
\newblock In {\em Microarchitecture (MICRO), 2016 49th Annual IEEE/ACM
  International Symposium on}, pages 1--12. IEEE, 2016.

\bibitem{zhou2016dorefa}
S.~Zhou, Y.~Wu, Z.~Ni, X.~Zhou, H.~Wen, and Y.~Zou.
\newblock Dorefa-net: Training low bitwidth convolutional neural networks with
  low bitwidth gradients.
\newblock {\em arXiv preprint arXiv:1606.06160}, 2016.

\bibitem{zhu2016trained}
C.~Zhu, S.~Han, H.~Mao, and W.~J. Dally.
\newblock Trained ternary quantization.
\newblock {\em arXiv preprint arXiv:1612.01064}, 2016.

\end{thebibliography}

\end{document}